\documentclass[letterpaper, 10 pt, journal, twoside]{IEEEtran}
\usepackage{amsmath,amsfonts}
\usepackage{algorithmic}
\usepackage{algorithm}
\usepackage{array}
\usepackage[caption=false,font=normalsize,labelfont=sf,textfont=sf]{subfig}
\usepackage{textcomp}
\usepackage{stfloats}
\usepackage{url}
\usepackage{verbatim}
\usepackage{graphicx}
\usepackage{cite}
\usepackage{CJKutf8}
\usepackage{xcolor}
\usepackage{multirow}
\usepackage{makecell}
\usepackage{hyperref}
\begin{document}
\begin{CJK}{UTF8}{gbsn}

\title{Whole-Body Constrained Learning for Legged Locomotion via Hierarchical Optimization}

\IEEEoverridecommandlockouts
\IEEEaftertitletext{\vspace{-1.5\baselineskip}}

\author{
Haoyu Wang, Ruyi Zhou$^{*}$, Liang Ding$^{*}$, \textit{Senior Member, IEEE}, Tie Liu, Zhelin Zhang, \\Peng Xu, Haibo Gao and Zongquan Deng 
\thanks{The authors are with the State Key Laboratory of Robotics and Systems, Harbin Institute of Technology, Harbin 150001, China.}
\thanks{$^*$Corresponding author: Ruyi Zhou (zhouryhit@gmail.com), Liang Ding (liangding@hit.edu.cn).}
\thanks{\textcolor{black}{Videos materials: 
\href{https://qrpucp.github.io/whole_body_constrained_learning/}{https://qrpucp.github.io/whole\_body\_constrained\_learning/}.}}
}

\maketitle
\begin{abstract}
Reinforcement learning (RL) has demonstrated impressive performance in legged locomotion over various challenging environments. However, due to the sim-to-real gap and lack of explainability, unconstrained RL policies deployed in the real world still suffer from inevitable safety issues, such as joint collisions, excessive torque, or foot slippage in low-friction environments. These problems limit its usage in missions with strict safety requirements, such as planetary exploration, nuclear facility inspection, and deep-sea operations.
In this paper, we design a hierarchical optimization-based whole-body follower, which integrates both hard and soft constraints into RL framework to make the robot move with better \textcolor{black}{safety} guarantees. 
Leveraging the advantages of model-based control, our approach allows for the definition of various types of hard and soft constraints during training or deployment, which allows for policy fine-tuning and mitigates the challenges of sim-to-real transfer. Meanwhile, it preserves the robustness of RL when dealing with locomotion in complex unstructured environments. The trained policy with introduced constraints was deployed in a hexapod robot and tested in various outdoor environments, including snow-covered slopes and stairs, demonstrating the great traversability and safety of our approach. 
\end{abstract}


\vspace{-0.2cm}
\section{INTRODUCTION}

Legged locomotion control is a challenging problem involving multi-body dynamics and interactions with diverse environments.
Conventional model-based approaches improved dynamic locomotion performance through precise modeling and optimal control techniques \cite{pratt2001virtual, carlo2018dynamic, bellicoso2016perception, Kim2019highly, shamel2020stance, nguyen2019optimized, katz2019minicheetah}, but they have limited adaptability to complex unstructured environments. Recently, RL-based methods trained in massive parallel simulations significantly advanced legged locomotion control \cite{joonho2020learning, takahiro2022learning, suyoung2023learning, jemin2019learning, liu2024visual, dongho2023rlmodel, fulong2021runlike, fuchioka2023optmimic, zhang2024imitation, kumar2021rma}, showcasing remarkable robustness and agility in complex and varied scenarios. However, policies trained in simulation still raise several \textcolor{black}{safety} concerns when directly deployed on physical robots in real-world environments. On the one hand, due to the sim-to-real gap \cite{kumar2021rma} regarding robot and terrain (discrepancies between simulation model/terrain and the real robot/terrain) and physics (inaccuracies in the simulation of collisions and contacts to various physical environments), the locomotion policies learned in simulation may exhibit different behavior when applied to the real world, potentially leading to joint collisions or excessive torque output. On the other hand, RL models lack interpretability and it is hard to guarantee their output satisfies certain physical constraints, especially when robots encounter conditions outside their training domain.
These safety concerns hinder the deployment of RL-based locomotion in safety-critical missions, where failure or unreliable movements could lead to serious risks to human safety, or damage to costly equipment.

Several approaches were proposed to address these safety concerns by reducing sim-to-real gap from different aspects, such as domain randomization \cite{xie2021dynamics}, simulator accuracy improvement \cite{suyoung2023learning}, actuator modeling \cite{hwangbo2019learning}, and privileged learning \cite{kumar2021rma}. 
Besides, some researchers have attempted to enhance the safety of RL policies. \cite{tairan2024agile, fan2024learn} trained recovery policies alongside locomotion policies, allowing automatic switching to safe policies in risky situations. Others employed constrained RL algorithms \cite{chane2024cat, lee2023evaluation, kim2024notonly} to ensure compliance with manually defined safety-related constraints. Since these methods are usually developed in simulation with the aim of transferring to real-world applications, constraint violation issues from sim-to-real mismatch will occur unpredictably during deployment. To overcome these issues, robotic researchers usually need to repetitively fine-tune in the simulation with intuitive judgment according to locomotion behavior observed during deployment, which is time-consuming and arduous.

In contrast, model-based control methods are less impacted by the sim-to-real gap and have higher levels of \textcolor{black}{safety} guarantees \cite{sehoon2025learning}. Their key advantage lies in the ability to formulate optimization problems that seamlessly integrate constraints into real robot control strategies \cite{carlo2018dynamic, Kim2019highly, shamel2020stance}, effectively regulating controller behavior in real-world scenarios. Additionally, they allow for straightforward adjustments of physically meaningful control parameters during deployment.

\begin{figure*}[hpt!]
    \centering
    \includegraphics[width = 1.0\textwidth]{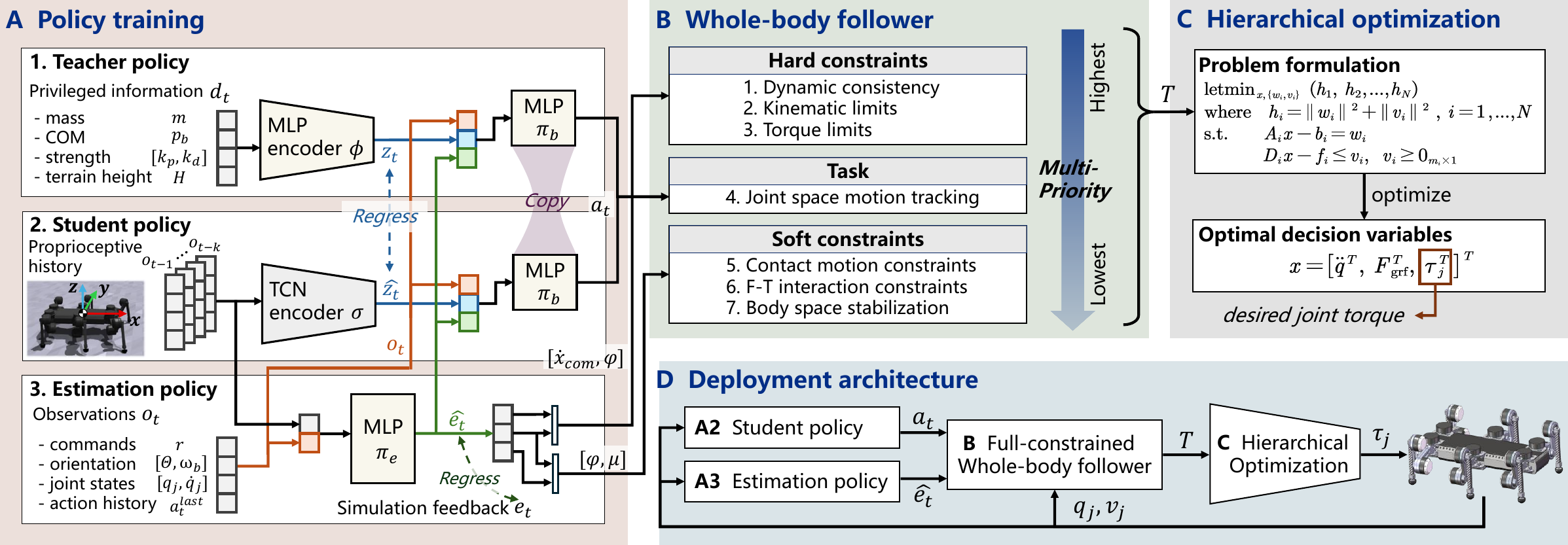}
    \vspace{-15pt}
    \caption{Overview of the presented whole-body constrained learning framework. 
    (A) Teacher-Student RL module.
    (B) Whole-body follower.
    (C) Hierarchical optimization.
    (D) Deployment architecture.
    }
    \vspace{-9pt}
    \label{fig:control_framework}
    \vspace{-9pt}
\end{figure*}

To enhance the safety of RL policies in real-world deployment and promote the application in safety-critical scenarios, we propose a whole-body constrained learning framework for legged locomotion, integrating the agility of learning-based methods with the reliability of model-based control, as illustrated in Fig.~\ref{fig:control_framework}. The framework employs a hierarchical optimization algorithm to manage multi-priority tasks through a whole-body follower. \textcolor{black}{It tracks the joint trajectories generated by RL policy while incorporating both hard and soft constraints. During training, only essential hard constraint like torque limits are enforced, relying on RL rewards to guide behavior.} At deployment, additional soft constraints such as foot-terrain interactions are added to enhance safety. Furthermore, an estimation policy is trained via supervised learning to estimate ground friction in real-time, enabling dynamic adjustment of constraint parameters during deployment, which facilitates improved adaptation to environmental conditions.

The main contributions of this work are summarized below:
\begin{itemize}
\item{A constrained RL framework is proposed, utilizing a whole-body follower based on hierarchical optimization as the low-level controller. This framework integrates both hard and soft constraints to enhance the safety of control policies while maintaining the robustness of RL.}
\item{By employing an optimization-based approach, constraints can be defined separately during training and deployment. This allows fine-tuning of the policy during deployment to better adapt to the environment, thereby reducing the difficulties of sim-to-real transfer.}
\item{Real-world locomotion experiments with a hexapod robot in various outdoor terrains demonstrating the robustness of trained policy, and the comparison with whole-body control and unconstrained RL, along with the constraints modification experiments, illustrate the effectiveness of our framework in enhancing safety and policy fine-tuning.}
\end{itemize}

\vspace{-0.15cm}
\section{RELATED WORK}

\subsection{Model-based Legged Locomotion}

Model-based methods rely on specific dynamic models to provide system inputs using manually designed control laws. Classical approaches often utilize simplified models, such as the Linear Inverted Pendulum (LIP) or Spring-Loaded Inverted Pendulum (SLIP), to design virtual models for controlling biped robots \cite{pratt2001virtual}. Model Predictive Control (MPC) solves the optimal control problem within a receding time horizon. Its predictive capability is especially critical for legged locomotion that includes a flight phase\cite{carlo2018dynamic}. Whole-Body Control (WBC) is a multi-priority controller that considers floating-base dynamics to compute desired joint torques while satisfying multi-priority task constraints\cite{bellicoso2016perception, Kim2019highly, shamel2020stance}. Another model-based control approach is trajectory optimization, which is typically used to generate specific maneuvers for legged robots, such as jumping\cite{nguyen2019optimized} and backflips\cite{katz2019minicheetah}.

These model-based methods often require expert knowledge and precise dynamic models but allow easy constraint integration and parameter adjustments based on actual performance, facilitating desired motions.

\vspace{-0.15cm}
\subsection{Learning-based Legged Locomotion}

Learning-based dynamic control methods have proven their effectiveness for legged robots\cite{jemin2019learning, joonho2020learning, takahiro2022learning}. Compared to model-based approaches, learning-based methods can handle more challenging terrains that are difficult to perceive and model, such as soft ground \cite{suyoung2023learning}, icy or snowy terrains \cite{takahiro2022learning}, and forests \cite{joonho2020learning}. 

Recently, some studies have explored incorporating model-based control methods into RL to enhance performance. One approach uses a hierarchical structure \cite{dongho2023rlmodel, fulong2021runlike}, where RL trains the high-level commands, which are then executed by model-based low-level controllers such as MPC. Another approach generates trajectories using trajectory optimization and then uses imitation learning to track them \cite{fuchioka2023optmimic}. Additionally, \cite{zhang2024imitation} pre-trains with model-based control, followed by imitation learning and fine-tuning with end-to-end RL.

The most comparable work, by Yin \textit{et al.} \cite{fulong2021runlike}, which also combines RL with whole-body methods. Their approach is a two-stage method that first uses RL to generate walking style parameters based on predefined reference contact states, and then employs splines to generate foot trajectories. In contrast, our method is a single-stage RL approach directly outputs joint trajectories, thereby fully leveraging RL’s capabilities.

\vspace{-0.2cm}
\subsection{Safe and Constrained Reinforcement Learning}

Learning-based locomotion control for legged robots rarely takes constraints into consideration. However, some recent research has shown that incorporating constraints or safety policies during training not only enhances safety but also significantly reduces the complexity of reward engineering \cite{kim2024notonly}. He \textit{et al.} trained recovery policies alongside locomotion policies to address safety violations \cite{tairan2024agile}.  Fan \textit{et al.} proposes replacing policy actions with safe actions when necessary to protect the learning agent and prevent constraint violations \cite{fan2024learn}. \cite{chane2024cat} introduces constraints by treating rewards as termination conditions. Additionally, some work apply constrained reinforcement learning algorithms to achieve robust locomotion in bipedal, quadrupedal, and wheeled robots \cite{lee2023evaluation, kim2024notonly}.

Compared to classical safe and constraint learning methods, our approach introduces constraints through hierarchical optimization, enabling distinct constraint settings during training and deployment. This allows for fine-tuning of policies during deployment by identifying or manually modifying parameters within predefined constraints.

\vspace{-0.1cm}
\section{MODEL FORMULATION}
\vspace{-0.1cm}

\begin{figure}[h]
 \vspace{-6pt}
    \centering
    \includegraphics[width = 0.48\textwidth]{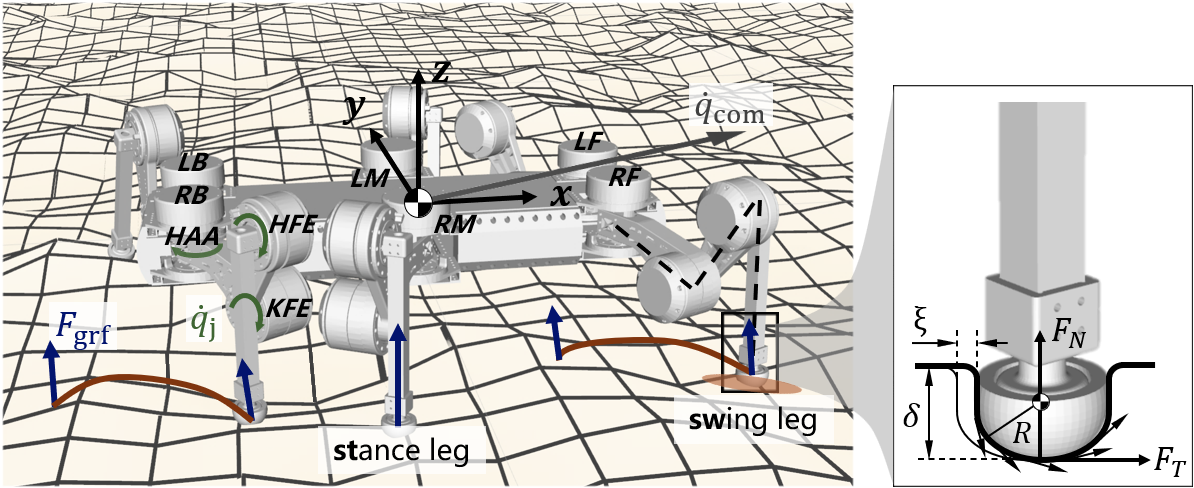}
    \vspace{-4pt}
    \caption{
    Illustration of the robotic dynamics model and the foot-terrain interaction model.
    }
    \vspace{-4pt}
    \label{fig:robot_model}
    \vspace{-2pt}
\end{figure}

As illustrated in Fig.~\ref{fig:robot_model}, for a legged robot with $n$ degrees of freedom and $c$ feet, assuming that all external forces act on the stance legs, the robot's dynamics can be expressed as:
\begin{equation}
\label{eq:dynamic}
\begin{gathered}
\underbrace{\left[\begin{array}{ccc}
M_{\mathrm{com}} & 0_{3 \times 3} & 0_{3 \times n} \\
0_{3 \times 3} & M_\theta & M_{\theta j} \\
0_{n \times 3} & M_{\theta j}^T & M_j
\end{array}\right]}_{\textcolor{black}{M}} \underbrace{\left[\begin{array}{c}
\ddot{x}_{\mathrm{com}} \\
\dot{\omega}_b \\
\ddot{q}_j
\end{array}\right]}_{\ddot{q}}+\underbrace{\left[\begin{array}{c}
h_{\mathrm{com}} \\
h_\theta \\
h_j
\end{array}\right]}_{\textcolor{black}{h}} \\
=\underbrace{\left[\begin{array}{c}
0_{3 \times 1} \\
0_{3 \times 1} \\
\tau_j
\end{array}\right]}_{\tau}+\underbrace{\left[\begin{array}{c}
J_{\mathrm{com}}^T \\
J_{\theta}^T \\
J_{j}^T
\end{array}\right]}_{\textcolor{black}{J^T}} F_{\mathrm{grf}},
\end{gathered}
\end{equation}
where the generalized robot state $ q = [x_{\mathrm{com}}^T, R_b^T, q_j^T]^T \in SE(3) \times \mathbb{R}^n $ combines the center of mass (CoM) position $ x_{\mathrm{com}} \in \mathbb{R}^3 $, body orientation $ R_b \in SO(3) $, and joint positions $ q_j \in \mathbb{R}^n $. Its derivatives yields the velocity state $ \dot{q} = [\dot{x}_{\mathrm{com}}^T, \omega_b^T, \dot{q}_j^T]^T \in \mathbb{R}^{6+n} $ and the acceleration state $ \ddot{q} = [\ddot{x}_{\mathrm{com}}^T, \dot{\omega}_b^T, \ddot{q}_j^T]^T \in \mathbb{R}^{6+n} $. The inertia matrix $ M \in \mathbb{R}^{(6+n) \times (6+n)} $ describes the system's inertia properties, while the nonlinear force vector $ h \in \mathbb{R}^{6+n} $ accounts for Coriolis, centrifugal, and gravitational forces. \textcolor{black}{The torque vector $\tau \in \mathbb{R}^{6+n}$ consists of a 6-dimensional uncontrollable torque and an $n$-dimensional actuated joint torque $\tau_j$. The ground reaction forces (GRFs) at the foot contacts are denoted by $ F_{\mathrm{grf}} \in \mathbb{R}^{n} $, and the Jacobian matrix for robot feet $ J \in \mathbb{R}^{n \times (6+n)} $ includes submatrices for CoM position $ J_{\mathrm{com}} \in \mathbb{R}^{n \times 3} $, body orientation $ J_{\theta} \in \mathbb{R}^{n \times 3} $, and joints $ J_{j} \in \mathbb{R}^{n \times n} $.}

In legged locomotion, foot-terrain interaction is crucial for ensuring the stability and efficiency of movement. To better adapt to various terrains, it is essential to model the foot-terrain interaction. In this work, we adopt the interaction model from \cite{ding2013foot}, which decomposes the interaction stress into normal bearing stress and tangential shear stress, using the stress integration format commonly employed in terramechanics. The normal force $ F_N $ and tangential force $ F_T $ exerted by the terrain on the foot are described as follows:
\begin{equation}
\textcolor{black}{
\begin{aligned}
\label{eq:interaction_model}
F_N&=\pi R^2\left(\frac{k_c}{R}+k_{\varphi}\right) \delta^m+b_N \dot{\delta}
\\
F_T&=\left(\pi R^2 a+ \mu F_N\right) \frac{1-e^{-1.43 \xi / K}}{1+e^{-1.43 \xi / K}}+b_T v\dot{\xi} ,
\end{aligned}}
\end{equation}
in which，$R$ denotes the radius of the foot. $k_c$, $k_{\varphi}$, $b_N$, $a$, $\mu$, $K$, and $b_T$ are respectively the cohesion modulus, internal friction modulus, normal damping coefficient, cohesion stress, friction coefficient, shear deformation modulus, and tangential damping coefficient of the soft terrain. $\delta$ represents the penetration depth of the foot. $m$ is the sinkage exponent, and $\xi$ represents the tangential slip displacement of the foot.

\vspace{-0.1cm}
\section{METHODOLOGY}

\subsection{Overview}

The overview of our framework is illustrated in Fig. \ref{fig:control_framework}. We use RL to generate desired joint trajectories and employ a constrained whole-body follower based on hierarchical optimization for tracking. Additionally, we concurrently train an estimation policy using supervised learning to provide observations for the whole-body follower. The deployment pipeline on the physical robot is shown in Fig. \ref{fig:control_framework}(D).

\vspace{-0.1cm}
\subsection{Whole-Body Control}

\textcolor{black}{
To enable accurate tracking of the joint trajectories generated by the RL policy while introducing constraints, we employ the Whole-Body Control (WBC) method. We formulate the WBC problem as a hierarchical quadratic program (HQP), solved using hierarchical optimization \cite{bellicoso2016perception}. Compared to projection-based WBC, the HQP-based method provides a more effective approach to handling multiple inequality constraints and has been widely adopted \cite{wensing2024optimization}. The decision variable of the HQP, denoted by $x \in \mathbb{R}^{6 + 3n}$, is defined as:}
\begin{equation}
\textcolor{black}{
    x=[\ddot{q}^{T}, F_{\text{grf}}^{T}, \tau_{j}^{T}]^T ,
}
\end{equation}
where $\ddot{q}$ represents the generalized accelerations, $F_{\text{grf}}$ denotes the ground reaction forces, and $\tau_j$ denotes the joint torques, as introduced in the system dynamics Eq. \eqref{eq:dynamic}.

Let $\text{TASK} = \{T_1, \ldots, T_N\}$ denote the set of WBC tasks, where $N$ is the total number of tasks. 
Each task $T_i$ is composed of equality and inequality constraints, given by:
\begin{equation}
  \label{eq:wbc_task}
  \textcolor{black}{
  {T_i}:\left\{\begin{array}{l}
  A_i {x}-{b_i}={w_i} \\
  {D_i x}-{f_i} \leq {v_i}
  \end{array}\right.,
  }
  \end{equation}
where $A_i \in \mathbb{R}^{n_i \times ({6+3n})}$, $b_i \in \mathbb{R}^{n_i}$, $D_i \in \mathbb{R}^{m_i \times ({6+3n})}$, $f_i \in \mathbb{R}^{m_i}$, $w_i \in \mathbb{R}^{n_i}$, and $v_i \in \mathbb{R}^{m_i}$ denote the task-specific matrices and vectors, and $w_i$ and $v_i$ denote the corresponding slack variables that should be minimized.

In our implementation, the WBC hierarchy includes three types of tasks: (i) physical constraint tasks $\{T_1, T_2, T_3\}$, (ii) the primary trajectory tracking task $T_4$, and (iii) performance-oriented tasks $\{T_5, T_6, T_7\}$. The detailed formulation of each task is presented below in descending order of priority.

\subsubsection{Dynamic Consistency}

To ensure consistency with the floating-base dynamics of the legged robot, as defined in Eq.~(\ref{eq:dynamic}), \textcolor{black}{we formulate the dynamic consistency task $T_1$ in the task space as:}
\begin{equation}
\textcolor{black}{
\underbrace{
\begin{bmatrix}
M & -J^T & -S_j
\end{bmatrix}
}_{A_1}
x
=
\underbrace{
- h
}_{b_1},}
\end{equation}
where \textcolor{black}{$S_j = [0_{n \times 6}, I_{n \times n}]^T$} denotes the joint selection matrix, which extracts the actuated joint torques $\tau_j$ from $\tau$.

\subsubsection{Kinematic Limits}

The kinematic limits task $T_2$ is designed to ensure that joint positions remain within predefined bounds $q_{j,\text{max}}$ and $q_{j,\text{min}}$ by imposing constraints in the acceleration space rather than directly on positions. These acceleration bounds are derived based on the current joint positions $q_j$, velocities $\dot{q}_j$, and a predefined deceleration duration $\Delta t$. 

Specifically, during the deceleration phase, each joint is required to reduce its velocity to zero while approaching the position limits, leading to the following expression for joint acceleration bounds:
\begin{equation}
    \textcolor{black}{
    \begin{aligned}
    \ddot{q}_{j,\text{max}}&=-2/\Delta t^2\left( q_{j,\text{max}}-q_j-\Delta t\dot{q}_j \right),
    \\
    \ddot{q}_{j,\text{min}}&=-2/\Delta t^2\left( q_{j,\text{min}}-q_j-\Delta t\dot{q}_j \right) ,
    \end{aligned}}
    \end{equation}
where $\Delta t = 10 \delta t$, and $\delta t$ denotes the control loop interval.
Given the constraint $\ddot{q}_{j,\text{min}} \leq \ddot{q}_j \leq \ddot{q}_{j,\text{max}}$, the corresponding inequality constraint in task space can be expressed as:
\begin{equation}
\textcolor{black}{
\underbrace{
\begin{bmatrix}
S_j^T & 0_{n \times 2n} \\
- S_j^T & 0_{n \times 2n}
\end{bmatrix}
}_{D_2}
x
\leq
\underbrace{
\begin{bmatrix}
\ddot{q}_{j,\text{max}} \\
\ddot{q}_{j,\text{min}}
\end{bmatrix}
}_{f_2}.}
\end{equation}

\subsubsection{Torque Limits}

The torque limits task $T_3$ ensures that the joint torques $\tau_j$ obtained from optimization remain within the allowable bounds imposed by the motors and actuators, and is represented as linear inequalities in the task space:
\begin{equation}
\underbrace{
\begin{bmatrix}
0_{n \times (2n + 6)} & I_{n \times n} \\
0_{n \times (2n + 6)} & -I_{n \times n}
\end{bmatrix}
}_{D_3}
x
\leq
\underbrace{
\begin{bmatrix}
\tau_{j,\text{max}} \\
\tau_{j,\text{min}}
\end{bmatrix}
}_{f_3},
\end{equation}
where $\tau_{j,\text{max}}$ and $\tau_{j,\text{min}}$ denote the upper and lower bounds of joint torques, respectively.

\subsubsection{Joint Space Motion Tracking}

The joint space motion tracking task $T_4$ is aims to make the robot follow the reference joint trajectories.
This task is defined in the acceleration space, where the desired joint acceleration $\ddot{q}_j^*$ is computed using a proportional-derivative (PD) controller:
\begin{equation}
\ddot{q}_j^* = k_p(a_t - q_j) - k_d \dot{q}_j,
\end{equation}
where $k_p$ and $k_d$ are the proportional (stiffness) and derivative (damping) gains, respectively, and $a_t$ denotes the reference joint position obtained from the RL policy.

To enforce tracking of the desired joint acceleration, an equality constraint is imposed in the task space:
\begin{equation}
\textcolor{black}{
\underbrace{
\begin{bmatrix}
S_j^T & 0_{n \times 2n}
\end{bmatrix}
}_{A_4}
x
=
\underbrace{
\ddot{q}_j^*
}_{b_4}.}
\end{equation}

\subsubsection{Contact Motion Constraints}

\textcolor{black}{The contact motion constraint task $T_5$} is introduced to prevent foot slippage by ensuring that the acceleration of the stance foot in the world coordinate frame remains zero. 
For any stance leg, the corresponding constraint is expressed as:
\begin{equation}
\textcolor{black}{
\underbrace{
\begin{bmatrix}
J & 0_{n \times 2n}
\end{bmatrix}
}_{A_5}
x
=
\underbrace{
- \dot{J} \, \dot{q}
}_{b_5}.}
\end{equation}

\subsubsection{Foot–terrain (F-T) interaction constraints}

To adhere to the interaction model defined in Eq. (\ref{eq:interaction_model}), we introduce \textcolor{black}{the foot-terrain interaction constraints Task $T_6$}.

We determine the maximum tangential slip displacement $\xi_{xy,\text{max}}$ in the $x$ and $y$ directions based on the foot's workspace and the desired control performance. The maximum penetration depth $\delta_{\text{max}}$ is then determined based on the motor's maximum output torque and the structural load-bearing limit, yielding the following:
\begin{equation}
\begin{aligned}
\label{eq:ft_task}
&F_{xy,\text{max}}=\frac{\sqrt{2}}{2}\left(\pi R^2 a+ \mu F_z\right) \frac{1-e^{-1.43 \xi_{xy,\text{max}} / K}}{1+e^{-1.43 \xi_{xy,\text{max}} / K}},
\\
&F_{z,\text{max}}=\left\{
\begin{aligned}
&(\pi R k_c + \pi R^2 k_{\varphi}) \delta^m_{\text{max}}, &\varphi_i = 1 \\ 
&0, &\varphi_i = 0 \\
\end{aligned}
\right.,
\end{aligned}
\end{equation}
in which \textcolor{black}{$\varphi \in \{0,1\}^c$} represents the ground contact state, where $\varphi_i=1$ indicates that leg $i$ is in contact with the ground, while \textcolor{black}{$\varphi_i = 0$} indicates that it is not in contact.

Transforming the constraints into the task space, we have:
\begin{equation}
\textcolor{black}{
\underbrace{
\begin{bmatrix}
0_{4c \times (n+6)} & S_f & 0_{4c \times n}
\end{bmatrix}
}_{D_6}
x
\leq
\underbrace{
f_f
}_{f_6},}
\end{equation}
with:
\begin{equation}
S_f = 
\begin{bmatrix}
S_0 & \dots & 0 \\
\vdots & \ddots & \vdots \\
0 & \dots & S_c
\end{bmatrix},\,\,\, 
\textcolor{black}{
f_f=
\begin{bmatrix}
f_1^T,\ldots,f_c^T
\end{bmatrix}^T,}
\end{equation}
in which, \textcolor{black}{$S_i\in\mathbb{R}^{4\times3},i = 0,\ldots,c$} and \textcolor{black}{$f_i \in \mathbb{R}^4,\,i = 0,\ldots,c$} are used to encode the foot–terrain interaction constraints from Eq. (\ref{eq:ft_task}) into the task space form presented in Eq. (\ref{eq:wbc_task}).

\subsubsection{Body Space Stabilization}

While tracking the joint trajectories generated by the RL policy, it is desirable to suppress undesired oscillations of the robot body, particularly in the roll, pitch, and vertical ($z$) directions. To this end, \textcolor{black}{the body space stabilization task $T_7$} is formulated to minimize the corresponding body accelerations through the following constraint:
\begin{equation}
\textcolor{black}{
\underbrace{
\begin{bmatrix}
S_b & 0_{n \times 2n}
\end{bmatrix}
}_{A_7}
x
=
\underbrace{
0_{n \times 1}
}_{b_7},}
\end{equation}
where $S_b \in \mathbb{R}^{n \times ({6+n})}$ denotes the body selection matrix, which is structurally analogous to the joint selection matrix $S_j$, and is used to extract the body acceleration components in the roll, pitch, and vertical directions. 

\begin{table}
\begin{center}
\vspace{-6pt}
\caption{Module Architecture Details}
\label{tab:module_architecture}
\vspace{-6pt}
\setlength{\tabcolsep}{1.0mm}{
\begin{tabular}{ccccc}
\hline
\textbf{Module} & \textbf{Inputs} & \textbf{Type} & \textbf{Hidden Layers} & \textbf{Outputs} \\ \hline
$\phi$ & $d_t$ & MLP & [128, 64, 20] & $z_t$ \\
$\sigma$ & $o_{[t-k:t-1]}$ & TCN & —— & $\hat{z_t}$ \\
$\pi_b$ & $o_t$, $z_t$/$\hat{z_t}$, $\hat{e_t}$ & MLP & [512, 256, 128] & $a_t$ \\
$\pi_e$ & $o_{[t-k:t]}$ & \textcolor{black}{TCN} & \textcolor{black}{——} & $\hat{e_t}$ \\ \hline
\end{tabular}}
\end{center}
\vspace{-18pt}
\end{table}

\vspace{-0.15cm}
\subsection{Reinforcement Learning}

\textcolor{black}{As shown in Fig.~\ref{fig:control_framework}(A), leveraging the Teacher-Student framework's implicit parameter identification capability\cite{joonho2020learning}, we train the RL policy using this method and employ the whole-body follower for trajectory tracking.} \textcolor{black}{Following the method from \cite{gwanghyeon2022concurrent}, the estimation policy is trained concurrently via supervised learning to provide observations for the whole-body follower.}

The network architecture, state-action space, reward definition, and the settings for whole-body follower are detailed in the following.

\subsubsection{Network architecture}

The network architecture is divided into three components:
\begin{itemize}
    \item \textbf{Teacher Policy} $\pi_t(o_t, d_t, \hat{e_t})$: Consists of an encoder $\phi$ and a body policy $\pi_b$. It leverages privileged information $d_t$ from simulation to learn advanced locomotion skills to traverse diverse terrains.
    \item \textbf{Student Policy} $\pi_s(o_t, o_{[t-k:t-1]}, \hat{e_t})$: Composed of an encoder $\sigma$ and the same body policy $\pi_b$, it uses $k$ steps of historical state information to approximate the feature vector $z_t$ encoded from privileged information, inheriting the locomotion skills acquired by the teacher policy.
    \item \textbf{Estimation Policy} $\pi_e(o_t, o_{[t-k:t-1]})$: Estimates body velocity, ground friction, and foot contact states, and provides observations for the WBC, replacing traditional Kalman filter-based state observers\cite{bloesch2013state}, \textcolor{black}{while enabling constraint adjustment based on environmental changes.}
\end{itemize}

Table \ref{tab:module_architecture} details the network structure and inputs/outputs of each module.

\subsubsection{Observation and action space}

The observation vector $o_t$ serves as input to the RL policy:
\begin{equation}
o_t = [\Theta,\omega_b,r,q_j,\dot{q}_j,a_t^{\text{last}}] \in \mathbb{R}^{10+3n},
\end{equation}
where $\Theta \in \mathbb{R}^4$ is the robot's orientation in quaternion form, $\omega_b \in \mathbb{R}^3$ is the IMU-reported angular velocity, $r \in \mathbb{R}^3$ contains user command velocities, $q_j, \dot{q}_j \in \mathbb{R}^n$ are joint positions and velocities, and $a_t^{\text{last}} \in \mathbb{R}^n$ is the previous action.

The privileged information vector $d_t$ inputs to encoder $\phi$:
\begin{equation}
d_t = [m, p_b, \mu, k_p, k_d, H] \in \mathbb{R}^{40+2n},
\end{equation}
including the robot's mass $m \in \mathbb{R}$, center of mass $p_b \in \mathbb{R}^3$, ground friction $\mu \in \mathbb{R}$, WBC stiffness and damping coefficients $k_p, k_d \in \mathbb{R}^n$, and terrain elevation map $H \in \mathbb{R}^{35}$ obtained from 35 scan points around the robot in the simulation environment.

The estimation policy $\pi_e$ outputs an estimate of the vector $e_t$, expressed as:
\begin{equation}
e_t = [\dot{x}_\text{com}, \varphi, \mu] \in \mathbb{R}^{4+c},
\end{equation}
where $\dot{x}_\text{com} \in \mathbb{R}^3$ is the body velocity, $\varphi \in \mathbb{R}^c$ indicates foot contact states, and $\mu \in \mathbb{R}$ is the ground friction coefficient.

The action vector $a_t \in \mathbb{R}^{n}$ output by the body policy $\pi_b$ represents the desired positions of the robot's joints.

\subsubsection{Reward definition}

\begin{table}
\begin{center}
\vspace{-6pt}
\caption{Reward Hyperparameters}
\label{tab:reward}
\vspace{-6pt}
\setlength{\tabcolsep}{0.9mm}{
\begin{tabular}{ccc}
\hline
    \textbf{Reward terms $r_i$} & \textbf{Expression $f_i$} & \textbf{Weight $w_i$} \\
\hline
Linear velocity tracking & $g(v_{b,xy}^{\text{cmd}}-v_{b,xy})$* & 5.5dt \\

Angular velocity tracking & $g(\omega_{b,z}^{\text{cmd}}-\omega_{b,z})$* & 3.5dt \\

Joint motion & $-\lvert\lvert\ddot{q}_j\lvert\lvert^2 +\lvert\lvert\dot{q}_j\lvert\lvert^2$ & 0.001dt \\

Energy penalty & $-\lvert\lvert \tau_j \lvert\lvert^2$ & 0.0008dt \\

Linear velocity penalty & $-\lvert\lvert v_{b,z} \lvert\lvert^2$ & 5.0dt \\

Action rate & $-\lvert\lvert \dot{a_t} \lvert\lvert^2$ & 0.001dt \\
 
Whole-body acceleration rate & $-\lvert\lvert \dddot{a_t} \lvert\lvert^2$ & 0.0002dt \\
\hline 
\end{tabular}}
\end{center}
* The expression for $g(x)$ is $\exp(-\lvert\lvert x \lvert\lvert^2 / 0.25)$.
\vspace{-12pt}
\end{table}

As shown in Table \ref{tab:reward}, to track the desired velocity commands while maintaining the stability of the robot body, we designed a series of rewards, and the total reward $r$ is a weighted sum of these specific reward terms as
\begin{equation}
r = \sum{r_i w_i}.
\end{equation}

The weights of the reward terms need to be adjusted through a curriculum approach, with final values in Table \ref{tab:reward}.

\subsubsection{Whole-body follower}

\begin{table}[]
    \centering
    \vspace{-6pt}
    \caption{Multiple Priority Tasks in Whole-Body Follower}
    \label{tab:wbc_tasks}
    \vspace{-6pt}
    \setlength{\tabcolsep}{1.0mm}{
    \begin{tabular}{cccc}
    \hline
    \textbf{Priority} & \textbf{Type} & \textbf{Task} & \textbf{Application} \\
    \hline
    \makecell{0} &
    \makecell{Equality\\Inequality\\Inequality} &
    \makecell{Dynamic consistency\\Torque limits\\Kinematic limits} &
    \makecell{Training\&\\Deployment} \\
    \hline
    \makecell{1} &
    \makecell{Equality} &
    \makecell{Joint space motion tracking} &
    \makecell{Training\&\\Deployment} \\
    \hline
    \makecell{2} &
    \makecell{Equality\\Inequality} &
    \makecell{Contact motion constraints\\F-T interaction constraints} &
    \makecell{Deployment} \\
    \hline
    3 & Equality & Body space stabilization & Deployment \\
    \hline
    \end{tabular}
    \vspace{-16pt}}
\end{table}

Based on whole-body control principles and the defined tasks aforementioned, Table \ref{tab:wbc_tasks} outlines the whole-body follower setup for training and deployment. \textcolor{black}{Priority 1 is the primary trajectory tracking task, constrained by priority 0 above and supplemented by soft constraints (priorities 2 and 3) below.}

During training, only hard constraints (e.g., torque and kinematic limits) are applied. Unlike \cite{chane2024cat}, which uses soft constraints like action rate limits, we rely solely on rewards and penalties to maintain the original style of RL. In deployment, we introduce adjustable constraints (e.g., F-T interaction), which can be tuned based on the robot's real-world performance or online parameter identification, eliminating the need for retraining the RL policy.

\vspace{-0.15cm}
\section{EXPERIMENTAL RESULTS}

\subsection{Implementation Details}

Our experiments are conducted on Elspider-Air, a hexapod robot platform with 18 degrees of freedom, featuring torque-controlled joints and weighing approximately 30 kg. The robot is equipped with a 6-axis IMU, an Intel NUC12 onboard computer, and foot-end pressure sensors.

For RL policy training, We parallelized the simulation environment to perform rollouts with 4096 environments simultaneously in Isaac Gym\cite{makoviychuk2021isaac}. The teacher and student policies are trained together using the Proximal Policy Optimization (PPO) algorithm \cite{schulman2017proximal} on an NVIDIA RTX 4070 GPU. 
Our curriculum design and domain randomization settings are similar to those in \cite{rudin2021learning}, with additional domain randomization applied to privileged information. To ensure the stability of the force-based whole-body follower, the simulation time step is set to 0.005 seconds.

\begin{figure}[h]
    \centering
    \includegraphics[width = 0.48\textwidth]{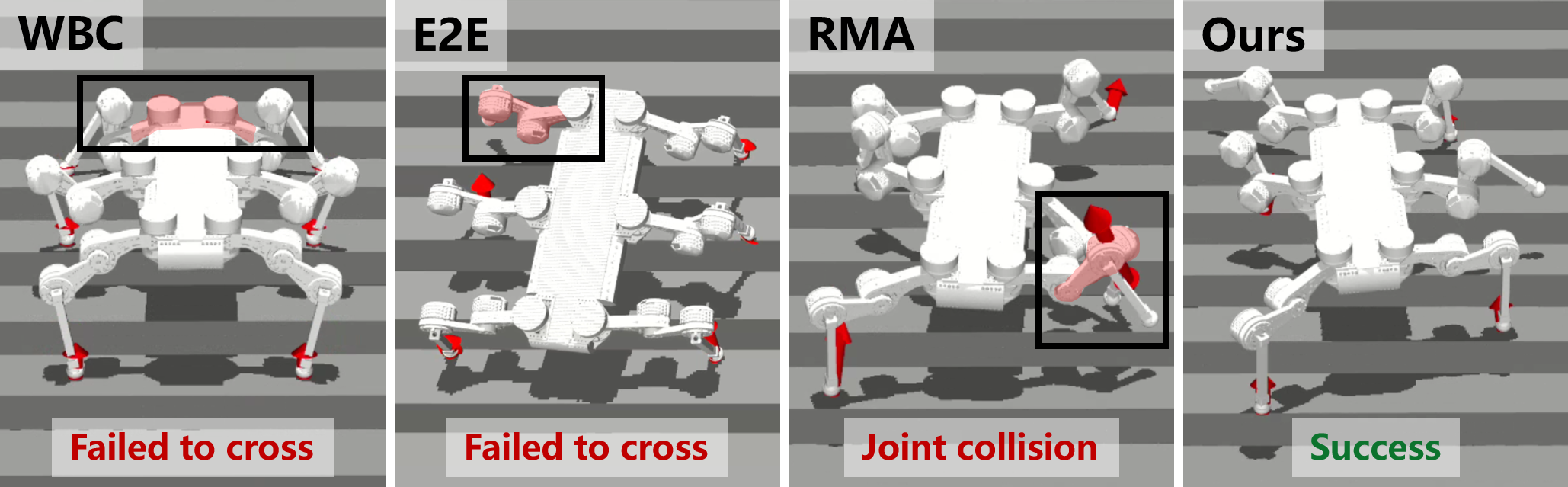}
    \vspace{-4pt}
    \caption{
    Performance comparison of stair climbing for WBC, RMA, End-to-End RL and our proposed method. Red arrows indicate the contact forces.
    }
    \vspace{-6pt}
    \label{fig:sim_comparison}
    \vspace{-4pt}
\end{figure}

For the whole-body follower, we utilize the Pinocchio dynamics library \cite{carpentier2019pinocchio} to solve floating-base dynamics. The hierarchical optimization algorithm is implemented by solving quadratic programming (QP) problems using qpOASES \cite{ferreau2014qpOASES}. During training, the whole-body follower operates at 50Hz, synchronized with the frequency of RL policy updates. In deployment, it runs at a higher frequency of 500 Hz to ensure real-time performance.

\vspace{-0.15cm}
\subsection{Simulation Experiments on Motion and Safety Performance}

To evaluate the motion-wise and safety-wise performance of our proposed framework, we conducted comparative experiments with other baselines in the MuJoCo\cite{todorov2012mujoco} simulation environment. The robot was tasked with traversing a 20-degree slope and 12 cm high stairs at a speed of 0.5 m/s, and its maximum achievable speed was also assessed. The settings for each baseline are detailed below:

\begin{itemize}
\item{\textbf{Whole-Body Controller (WBC)}: \textcolor{black}{A convex model predictive controller computes the target center of mass (CoM) acceleration\cite{carlo2018dynamic}, while the whole-body controller\cite{bellicoso2016perception} tracks it along with swing leg trajectories and ground reaction forces to generate joint torques. The same constraints as in our method are applied, but the friction value is predefined since no estimation policy is used.}}

\item{\textcolor{black}{\textbf{End-to-End RL (E2E)}: The simplest unconstrained end-to-end RL method, trained in Isaac Gym using the open-source Legged Gym framework \cite{rudin2021learning}.}}

\item{\textbf{RMA}: An unconstrained RL approach using the Teacher-Student method for rapid motor adaptation, trained in Isaac Gym, with the observation space, action space, and reward functions detailed in \cite{kumar2021rma}.}
\end{itemize}

\begin{table}[]
    \begin{center}
    \caption{Performance Comparison on Motion and Safety with Baseline Methods}
    \label{tab:outdoor_comparison}
    \vspace{-6pt}
    \setlength{\tabcolsep}{0.9mm}{
    \begin{tabular}{cccccccc}
    \hline
        \multirow{3}{*}{\textbf{\makecell{Control\\Method}}} &
        \multicolumn{4}{|c}{\textbf{Motion Performance}} &
        \multicolumn{3}{|c}{\textbf{Safety Performance}} \\
    \cline{2-8}
         &
        \multicolumn{2}{|c}{\textbf{Max Speed}} &
        \multicolumn{2}{c}{\textbf{Success Rate}} &
        \multicolumn{3}{|c}{\textbf{Hazardous Situations}} \\
    \cline{2-8}
         &
        \multicolumn{1}{|c}{linear} &
        \multicolumn{1}{c}{angular} &
        \multicolumn{1}{c}{slopes} &
        \multicolumn{1}{c}{stairs} &
        \multicolumn{1}{|c}{slippage} &
        \multicolumn{1}{c}{collision} &
        \multicolumn{1}{c}{torque} \\
    \hline
        WBC\cite{carlo2018dynamic}\cite{bellicoso2016perception} &
        \textcolor{black}{1.8 m/s} &
        \textcolor{black}{2.0 rad/s} &
        50\% &
        Failed &
        1 &
        1 &
        0 \\
        \textcolor{black}{E2E\cite{rudin2021learning}} &
        \textcolor{black}{2.2 m/s} &
        \textcolor{black}{2.5 rad/s} &
        \textcolor{black}{100\%} &
        \textcolor{black}{Failed} &
        \textcolor{black}{5} &
        \textcolor{black}{4} &
        \textcolor{black}{5} \\
        RMA\cite{kumar2021rma} &
        2.4 m/s&
        2.6 rad/s &
        100\% &
        100\% &
        2 &
        6 &
        6 \\
        \textbf{Ours} &
        2.3 m/s&
        2.6 rad/s &
        100\% &
        100\% &
        1 &
        2 &
        0 \\
    \hline
    \end{tabular}}
    \end{center}
    \vspace{-18pt}
\end{table}

To quantify the safety performance, we defined hazardous situations as foot slips more than 4 cm, joint torque exceeds 20 Nm, or joint collisions occur. For each method, we tested the robot's maximum speed on flat ground and conducted ten trials each on slopes and stairs, counting instances of hazardous situations (multiple counts per trial were possible). 
The experimental results are summarized in Table~\ref{tab:outdoor_comparison}.

From the Table~\ref{tab:outdoor_comparison} and Fig.~\ref{fig:sim_comparison}, it can be observed that the WBC method significantly decreases the frequency of hazardous situations by directly incorporating constraints through optimization. However, its reliance on expert-defined control laws limits movement speed \textcolor{black}{(max linear speed: 1.8 m/s, max angular speed: 2.0 rad/s) and traversability. In contrast, RL-based methods achieve faster speeds and better adaptation to challenging terrains, illustrated in higher success rate on slopes. Moreover, RMA excels on steps due to the implicit parameter identification capability of the Teacher-Student method. However, the absence of explicit constraints in RL-based methods results in a higher occurrence of hazardous situations.} 
Our proposed method effectively traverses terrains such as stairs and slopes, achieving linear speeds of up to 2.3 m/s and angular speeds of up to 2.6 rad/s, with only three instances of hazardous situations among twenty tests. The comprehensive motion-wise and safety-wise performance comparison suggests that our method preserves the intelligence and robustness of RL while mitigating unsafe occurrences through the inclusion of constraints.

\vspace{-0.15cm}
\subsection{Real-world Performance Analysis Experiments}

We further validate the performance of our proposed framework on a physical robot, evaluating it from motion and safety aspects, as shown below:

\subsubsection{Safety performance}

\textcolor{black}{Using the same RL policy, we compared the whole-body follower with a PD follower (classical unconstrained RL) in terms of tracking performance at a 1 m/s velocity command.}                          
To create a low-friction indoor environment, we covered the foam padding with oil-coated plastic film and wrapped the robot's feet in plastic sheeting to minimize ground friction.

\begin{figure}[h]
    \centering
    \includegraphics[width = 0.48\textwidth]{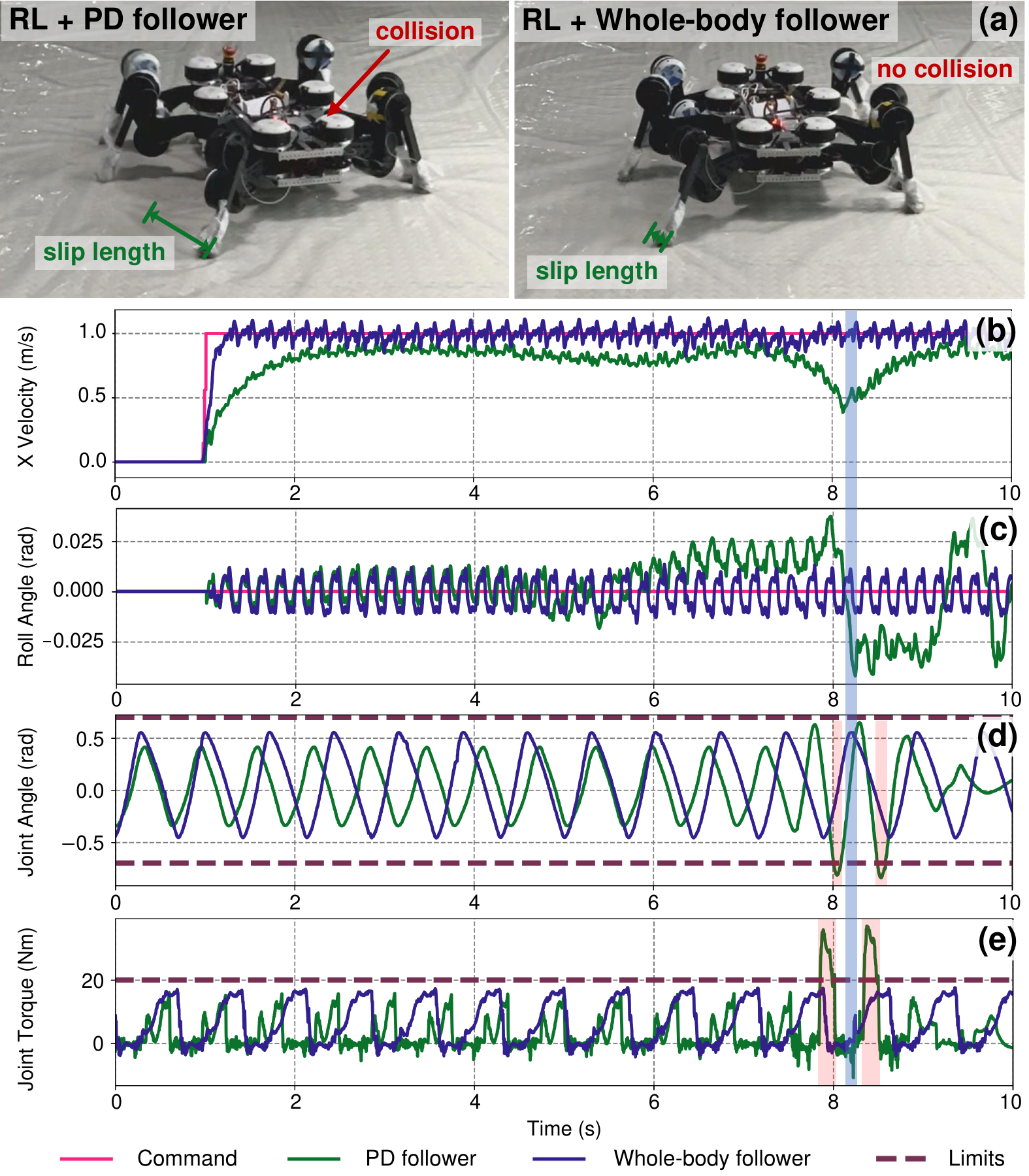}
    \vspace{-4pt}
    \caption{
    Locomotion performance comparison of PD follower and whole-body follower on a low-friction surface. (a) Slip length and collision comparison of two followers. (b-e) Robot's velocity in x-direction, body roll angle, joint angle, and joint torque results during the experiments. The blue region corresponds to the frozen frames of the robot at that moment, while the pink regions highlight areas where constraints are violated.
    }
    \vspace{-6pt}
    \label{fig:controller_comparison}
\end{figure}

As shown in Fig.~\ref{fig:controller_comparison}, the whole-body follower demonstrates superior performance on low-friction surfaces. It tracked the 1 m/s target velocity command within 0.3 seconds, with roll angles fluctuating by less than 0.03 radians and no joint collisions or torque limit exceedances occurred throughout the experiment. In contrast, the PD follower took 1 second to track the target velocity command and exhibited steady-state error. Additionally, severe slipping occurred around the 8-second mark (blue region), leading to torque limit exceedances and joint collision as highlighted in the pink regions of Fig.~\ref{fig:controller_comparison}.

\subsubsection{Motion performance}

\begin{figure*}[hpt!]
    \centering
    \includegraphics[width = 1.0\textwidth]{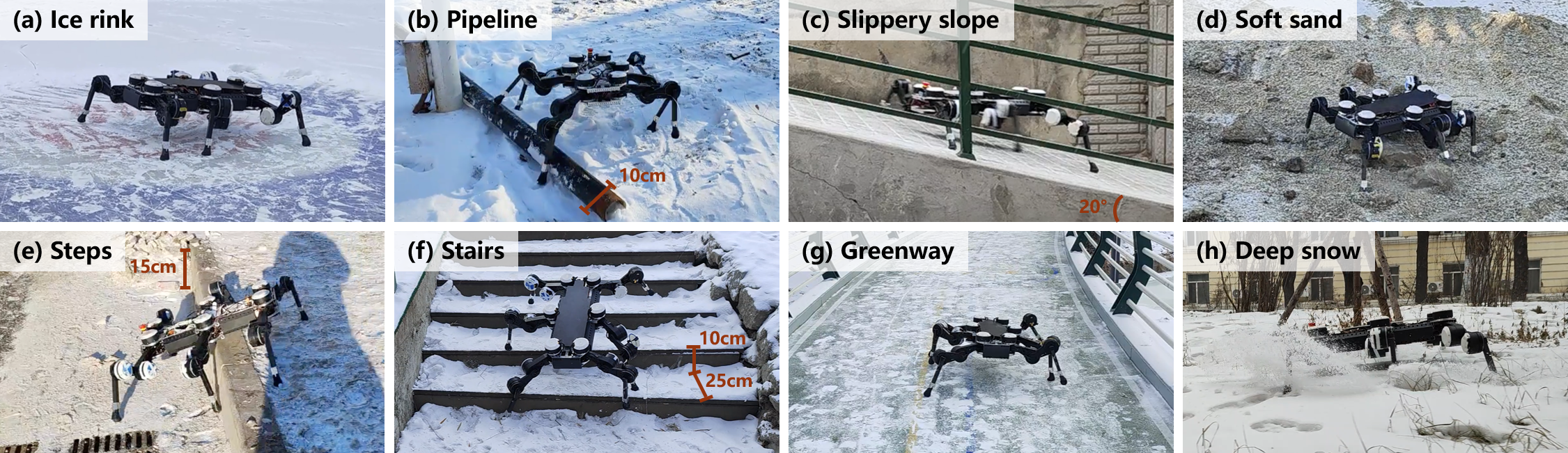}
    \vspace{-15pt}
    \caption{
    \textcolor{black}{Outdoor experiments on a variety of terrains, such as ice rink (a), pipeline (b), slope (c), sand (d), steps (e), stairs (f), greenway (g), and deep snow (h).} Our proposed framework exhibits excellent traversability and robustness, effectively navigating through diverse and challenging terrains while maintaining high levels of safety and reliability.
    }
    \vspace{-9pt}
    \label{fig:outdoor_experiment}
    \vspace{-9pt}
\end{figure*}

Smooth ice surfaces and soft snow present highly challenging terrains where walking can easily become hazardous even for humans. \textcolor{black}{To verify the motion performance of our proposed framework, we conducted multi-terrain walking experiments on snow and ice using the Elspider-Air robot, along with a comparative test against the whole-body controller.} The experimental results, as shown in Fig.~\ref{fig:outdoor_experiment} and the supplementary video, demonstrate that our method effectively adapts to complex terrains \textcolor{black}{such as icy rink, pipeline, slippery slope, soft sand, steps, stairs, greenway, and deep snow.}

\vspace{-0.15cm}
\subsection{Constraints Modification Experiments}

Our framework allows for defining constraints separately during training and deployment. In the following experiments, we analyze the impact of different constraint settings on policy deployment. Using the same RL policy with a target speed of 1 m/s, we employed a constrained whole-body follower for trajectory tracking while selectively modifying specific constraint parameters.

\subsubsection{Foot–terrain interaction constraints}

\begin{figure}[h]
    \centering
    \includegraphics[width = 0.48\textwidth]{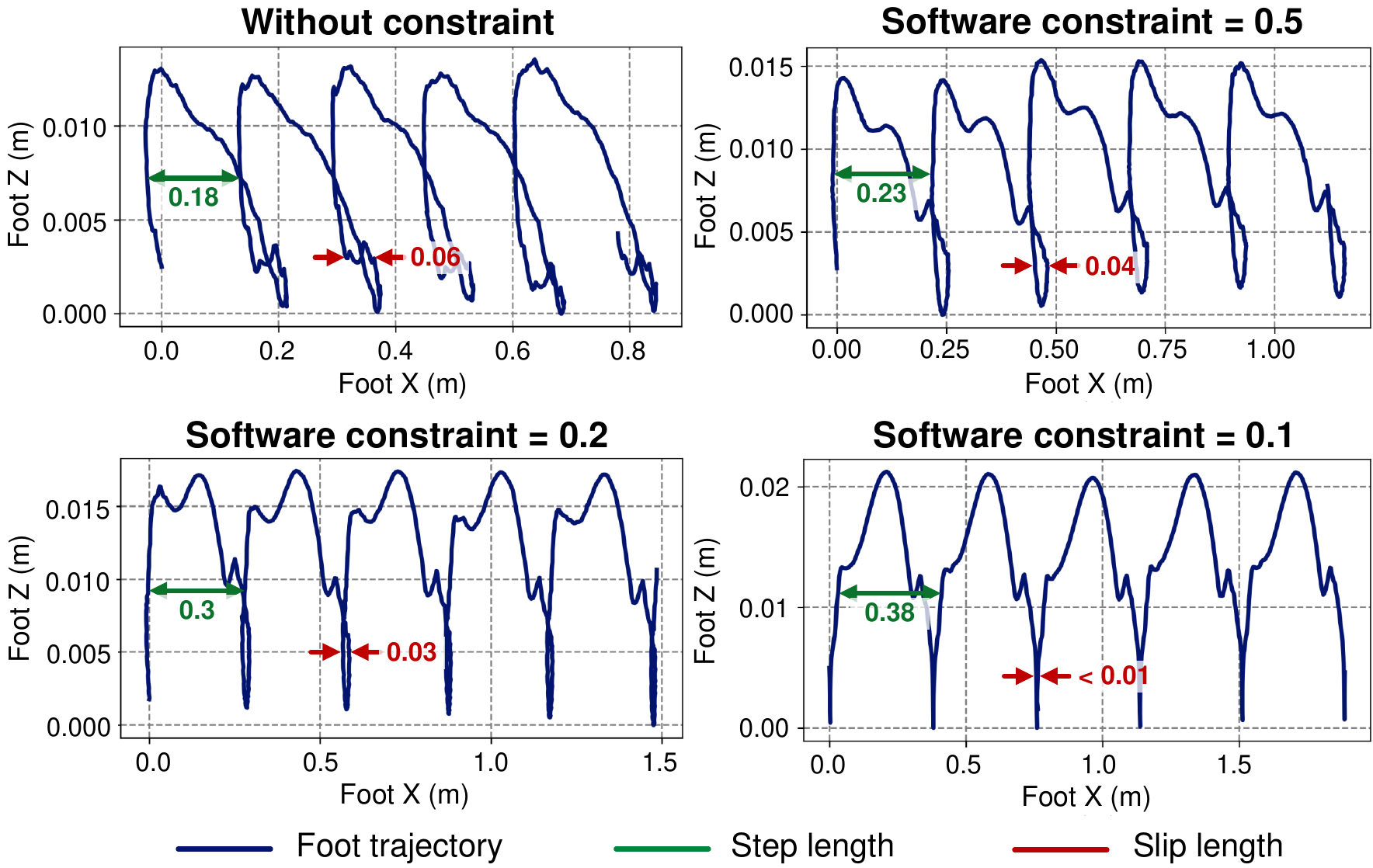}
    \vspace{-4pt}
    \caption{
    The trajectory of the right foreleg's foot after swinging five times with different foot-terrain interaction constraints set during deployment.
    }
    \vspace{-6pt}
    \label{fig:friction_constraint}
\end{figure}

We modified the foot-terrain interaction constraints by fixing the friction coefficient $\mu$ to 0.5, 0.2, and 0.1, instead of using the estimated values, and conducted comparative experiments on the aforementioned low-friction surface.

As shown in Fig.~\ref{fig:friction_constraint}, without foot–terrain interaction constraints, the foot slip was 6 mm, and the effective step length per stride was only 0.18 m. In contrast, with the introduction of foot–terrain interaction constraints, reducing the friction coefficient $\mu$ led to less foot slip and a longer effective step length. This demonstrates that the constraint settings directly influence the robot's slippage on the low-friction surface.

\subsubsection{Kinematic limits}

We modified the kinematic limits by limiting the range of motion (ROM) of the hip joint to 40°, 25°, and 15°, then compared these settings against a scenario with no kinematic constraints. 

\begin{figure}[h]
    \centering
    \includegraphics[width = 0.48\textwidth]{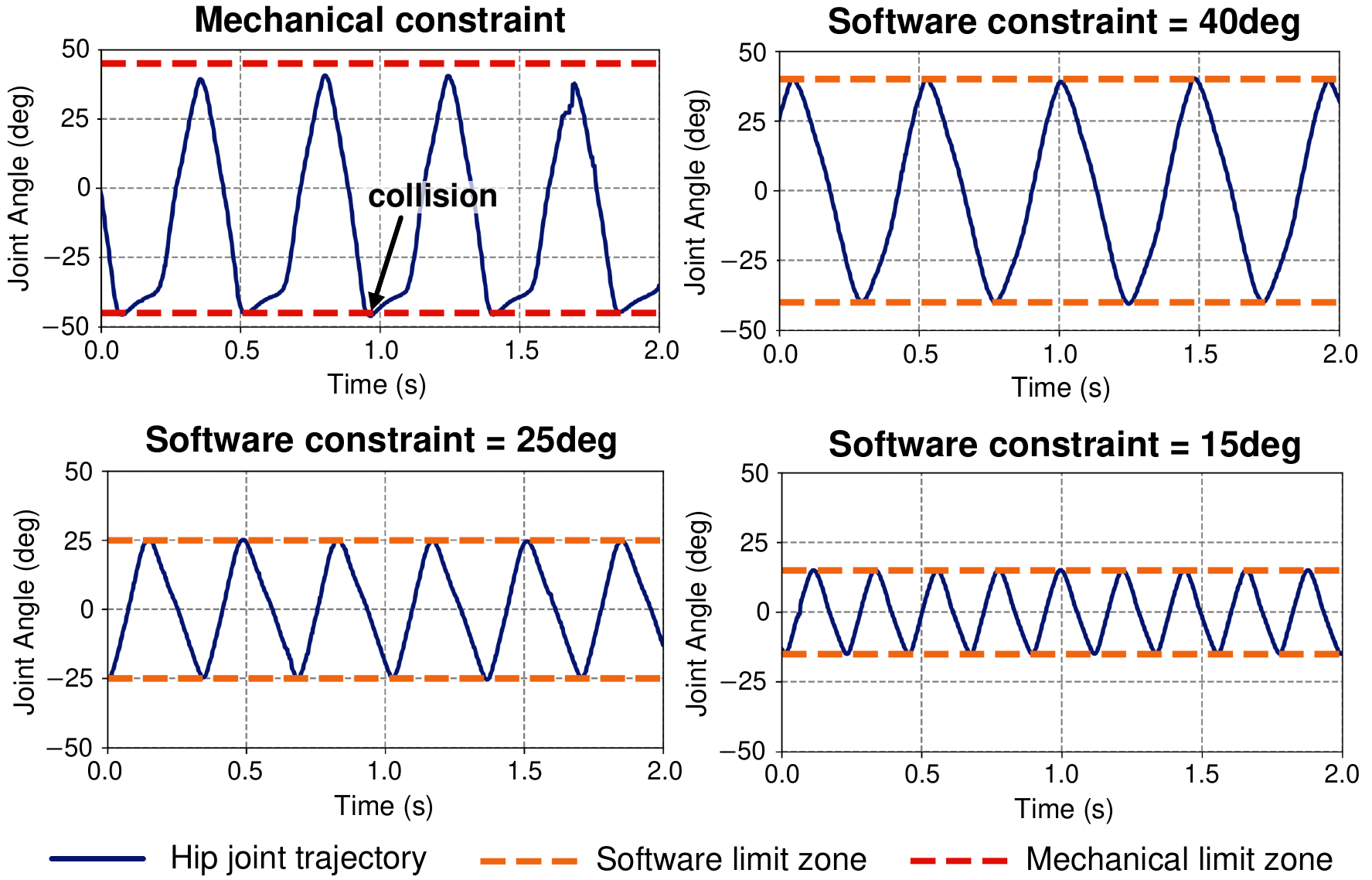}
    \vspace{-4pt}
    \caption{
    The trajectory of the right foreleg's hip joint within 2 seconds with different kinematic limits set during deployment.
    }
    \vspace{-6pt}
    \label{fig:position_constraint}
\end{figure}

As shown in Fig.~\ref{fig:position_constraint}, it is evident that the motion of the hip joint is confined within the specified constraint ranges. In contrast, without any constraints, the hip joint collided with its mechanical limits, significantly affecting the joint trajectory.

We also observed that as the hip joint's ROM decreases, its swing frequency must increase to maintain the same velocity. However, if the ROM is too small, the robot cannot achieve the desired velocity due to the maximum joint velocity limit.

The above experiments clearly show that modifying constraints during deployment can significantly impact performance. By adjusting constraints through estimation policy or manual parameter tuning, we can adapt the policy's behavior to better suit the environment, thereby reducing the challenges of sim-to-real transfer, all without altering the original RL policy. \textcolor{black}{However, these adjustments may also influence the original style of the RL policy to some extent. Therefore, a trade-off is required.}

\vspace{-0.15cm}
\section{CONCLUSION}

In this paper, we propose a constrained RL framework that uses a hierarchical optimization-based whole-body follower as the low-level controller, which enables flexible definitions of hard and soft constraints during training and deployment. 
Simulation experiments and hardware performance analysis experiments show that our method significantly improves RL policy safety while maintaining the high traversability and robustness of RL. 
Constraints modification experiments reveal that adjusting constraints during deployment aids policy fine-tuning and accelerates sim-to-real transfer. While demonstrated on a hexapod robot, the framework is adaptable to other legged robots with similar configurations.

In our future work, we plan to dynamically identify additional terrain parameters, which would further enhance the robot's adaptability to the environment.

\vspace{-0.1cm}
\bibliographystyle{IEEEtran}
\bibliography{IEEEexample}

\begin{thebibliography}{10}
\providecommand{\url}[1]{#1}
\csname url@rmstyle\endcsname
\providecommand{\newblock}{\relax}
\providecommand{\bibinfo}[2]{#2}
\providecommand\BIBentrySTDinterwordspacing{\spaceskip=0pt\relax}
\providecommand\BIBentryALTinterwordstretchfactor{4}
\providecommand\BIBentryALTinterwordspacing{\spaceskip=\fontdimen2\font plus
\BIBentryALTinterwordstretchfactor\fontdimen3\font minus \fontdimen4\font\relax}
\providecommand\BIBforeignlanguage[2]{{%
\expandafter\ifx\csname l@#1\endcsname\relax
\typeout{** WARNING: IEEEtran.bst: No hyphenation pattern has been}%
\typeout{** loaded for the language `#1'. Using the pattern for}%
\typeout{** the default language instead.}%
\else
\language=\csname l@#1\endcsname
\fi
#2}}

\bibitem{pratt2001virtual}
J.~Pratt, C.-M. Chew, A.~Torres, P.~Dilworth, and G.~Pratt, ``Virtual model control: An intuitive approach for bipedal locomotion,'' \emph{The International Journal of Robotics Research}, vol.~20, no.~2, pp. 129--143, 2001.

\bibitem{carlo2018dynamic}
J.~Di~Carlo, P.~M. Wensing, B.~Katz, G.~Bledt, and S.~Kim, ``Dynamic locomotion in the mit cheetah 3 through convex model-predictive control,'' in \emph{2018 IEEE/RSJ International Conference on Intelligent Robots and Systems (IROS)}, 2018, pp. 1--9.

\bibitem{bellicoso2016perception}
C.~Dario~Bellicoso, C.~Gehring, J.~Hwangbo, P.~Fankhauser, and M.~Hutter, ``Perception-less terrain adaptation through whole body control and hierarchical optimization,'' in \emph{2016 IEEE-RAS 16th International Conference on Humanoid Robots (Humanoids)}, 2016, pp. 558--564.

\bibitem{Kim2019highly}
\BIBentryALTinterwordspacing
D.~Kim, J.~D. Carlo, B.~Katz, G.~Bledt, and S.~Kim, ``Highly dynamic quadruped locomotion via whole-body impulse control and model predictive control,'' \emph{ArXiv}, vol. abs/1909.06586, 2019. [Online]. Available: \url{https://api.semanticscholar.org/CorpusID:202577921}
\BIBentrySTDinterwordspacing

\bibitem{shamel2020stance}
S.~Fahmi, M.~Focchi, A.~Radulescu, G.~Fink, V.~Barasuol, and C.~Semini, ``Stance: Locomotion adaptation over soft terrain,'' \emph{IEEE Transactions on Robotics}, vol.~36, no.~2, pp. 443--457, 2020.

\bibitem{nguyen2019optimized}
Q.~Nguyen, M.~J. Powell, B.~Katz, J.~D. Carlo, and S.~Kim, ``Optimized jumping on the mit cheetah 3 robot,'' in \emph{2019 International Conference on Robotics and Automation (ICRA)}, 2019, pp. 7448--7454.

\bibitem{katz2019minicheetah}
B.~Katz, J.~D. Carlo, and S.~Kim, ``Mini cheetah: A platform for pushing the limits of dynamic quadruped control,'' in \emph{2019 International Conference on Robotics and Automation (ICRA)}, 2019, pp. 6295--6301.

\bibitem{joonho2020learning}
J.~Lee, J.~Hwangbo, L.~Wellhausen, V.~Koltun, and M.~Hutter, ``Learning quadrupedal locomotion over challenging terrain,'' \emph{Science Robotics}, vol.~5, no.~47, p. eabc5986, 2020.

\bibitem{takahiro2022learning}
T.~Miki, J.~Lee, J.~Hwangbo, L.~Wellhausen, V.~Koltun, and M.~Hutter, ``Learning robust perceptive locomotion for quadrupedal robots in the wild,'' \emph{Science Robotics}, vol.~7, no.~62, p. eabk2822, 2022.

\bibitem{suyoung2023learning}
S.~Choi, G.~Ji, J.~Park, H.~Kim, J.~Mun, J.~H. Lee, and J.~Hwangbo, ``Learning quadrupedal locomotion on deformable terrain,'' \emph{Science Robotics}, vol.~8, no.~74, p. eade2256, 2023.

\bibitem{jemin2019learning}
J.~Hwangbo, J.~Lee, A.~Dosovitskiy, D.~Bellicoso, V.~Tsounis, V.~Koltun, and M.~Hutter, ``Learning agile and dynamic motor skills for legged robots,'' \emph{Science Robotics}, vol.~4, no.~26, p. eaau5872, 2019.

\bibitem{liu2024visual}
M.~Liu, Z.~Chen, X.~Cheng, Y.~Ji, R.-Z. Qiu, R.~Yang, and X.~Wang, ``Visual whole-body control for legged loco-manipulation,'' in \emph{8th Annual Conference on Robot Learning}, 2024.

\bibitem{dongho2023rlmodel}
D.~Kang, J.~Cheng, M.~Zamora, F.~Zargarbashi, and S.~Coros, ``Rl + model-based control: Using on-demand optimal control to learn versatile legged locomotion,'' \emph{IEEE Robotics and Automation Letters}, vol.~8, no.~10, pp. 6619--6626, 2023.

\bibitem{fulong2021runlike}
F.~Yin, A.~Tang, L.~Xu, Y.~Cao, Y.~Zheng, Z.~Zhang, and X.~Chen, ``Run like a dog: Learning based whole-body control framework for quadruped gait style transfer,'' in \emph{2021 IEEE/RSJ International Conference on Intelligent Robots and Systems (IROS)}, 2021, pp. 8508--8514.

\bibitem{fuchioka2023optmimic}
Y.~Fuchioka, Z.~Xie, and M.~Van~de Panne, ``Opt-mimic: Imitation of optimized trajectories for dynamic quadruped behaviors,'' in \emph{2023 IEEE International Conference on Robotics and Automation (ICRA)}, 2023, pp. 5092--5098.

\bibitem{zhang2024imitation}
Z.~Zhang, T.~Liu, L.~Ding, H.~Wang, P.~Xu, H.~Yang, H.~Gao, Z.~Deng, and J.~Pajarinen, ``Imitation-enhanced reinforcement learning with privileged smooth transition for hexapod locomotion,'' \emph{IEEE Robotics and Automation Letters}, vol.~10, no.~1, pp. 350--357, 2025.

\bibitem{kumar2021rma}
A.~Kumar, Z.~Fu, D.~Pathak, and J.~Malik, ``{RMA: Rapid motor adaptation for legged robots},'' in \emph{Robotics: Science and Systems}, 2021.

\bibitem{xie2021dynamics}
Z.~Xie, X.~Da, M.~van~de Panne, B.~Babich, and A.~Garg, ``Dynamics randomization revisited: A case study for quadrupedal locomotion,'' in \emph{2021 IEEE International Conference on Robotics and Automation (ICRA)}, 2021, pp. 4955--4961.

\bibitem{hwangbo2019learning}
J.~Hwangbo, J.~Lee, A.~Dosovitskiy, D.~Bellicoso, V.~Tsounis, V.~Koltun, and M.~Hutter, ``Learning agile and dynamic motor skills for legged robots,'' \emph{Science Robotics}, vol.~4, no.~26, p. eaau5872, 2019.

\bibitem{tairan2024agile}
T.~He, C.~Zhang, W.~Xiao, G.~He, C.~Liu, and G.~Shi, ``Agile but safe: Learning collision-free high-speed legged locomotion,'' in \emph{Robotics: Science and Systems (RSS)}, 2024.

\bibitem{fan2024learn}
K.~Fan, Z.~Chen, G.~Ferrigno, and E.~D. Momi, ``Learn from safe experience: Safe reinforcement learning for task automation of surgical robot,'' \emph{IEEE Transactions on Artificial Intelligence}, vol.~5, no.~7, pp. 3374--3383, 2024.

\bibitem{chane2024cat}
E.~Chane-Sane, P.-A. Leziart, T.~Flayols, O.~Stasse, P.~Sou{\`e}res, and N.~Mansard, ``Cat: Constraints as terminations for legged locomotion reinforcement learning,'' in \emph{IEEE/RSJ International Conference on Intelligent Robots and Systems (IROS)}, 2024.

\bibitem{lee2023evaluation}
\BIBentryALTinterwordspacing
J.~Lee, L.~Schroth, V.~Klemm, M.~Bjelonic, A.~Reske, and M.~Hutter, ``Evaluation of constrained reinforcement learning algorithms for legged locomotion,'' 2023. [Online]. Available: \url{https://arxiv.org/abs/2309.15430}
\BIBentrySTDinterwordspacing

\bibitem{kim2024notonly}
Y.~Kim, H.~Oh, J.~Lee, J.~Choi, G.~Ji, M.~Jung, D.~Youm, and J.~Hwangbo, ``Not only rewards but also constraints: Applications on legged robot locomotion,'' \emph{IEEE Transactions on Robotics}, vol.~40, pp. 2984--3003, 2024.

\bibitem{sehoon2025learning}
S.~Ha, J.~Lee, M.~Panne, Z.~Xie, W.~Yu, and M.~Khadiv, ``Learning-based legged locomotion: State of the art and future perspectives,'' \emph{The International Journal of Robotics Research}, 01 2025.

\bibitem{ding2013foot}
L.~Ding, H.~Gao, Z.~Deng, J.~Song, Y.~Liu, G.~Liu, and K.~Iagnemma, ``Foot–terrain interaction mechanics for legged robots: Modeling and experimental validation,'' \emph{The International Journal of Robotics Research}, vol.~32, no.~13, pp. 1585--1606, 2013.

\bibitem{wensing2024optimization}
P.~M. Wensing, M.~Posa, Y.~Hu, A.~Escande, N.~Mansard, and A.~D. Prete, ``Optimization-based control for dynamic legged robots,'' \emph{IEEE Transactions on Robotics}, vol.~40, pp. 43--63, 2024.

\bibitem{gwanghyeon2022concurrent}
G.~Ji, J.~Mun, H.~Kim, and J.~Hwangbo, ``Concurrent training of a control policy and a state estimator for dynamic and robust legged locomotion,'' \emph{IEEE Robotics and Automation Letters}, vol.~7, no.~2, pp. 4630--4637, 2022.

\bibitem{bloesch2013state}
M.~Bloesch, M.~Hutter, M.~A. Hoepflinger, S.~Leutenegger, C.~Gehring, C.~D. Remy, and R.~Siegwart, ``State estimation for legged robots: Consistent fusion of leg kinematics and imu,'' 2013.

\bibitem{makoviychuk2021isaac}
V.~Makoviychuk, L.~Wawrzyniak, Y.~Guo, M.~Lu, K.~Storey, M.~Macklin, D.~Hoeller, N.~Rudin, A.~Allshire, A.~Handa, and G.~State, ``Isaac gym: High performance gpu-based physics simulation for robot learning,'' 2021.

\bibitem{schulman2017proximal}
\BIBentryALTinterwordspacing
J.~Schulman, F.~Wolski, P.~Dhariwal, A.~Radford, and O.~Klimov, ``Proximal policy optimization algorithms,'' 2017. [Online]. Available: \url{https://arxiv.org/abs/1707.06347}
\BIBentrySTDinterwordspacing

\bibitem{rudin2021learning}
N.~Rudin, D.~Hoeller, P.~Reist, and M.~Hutter, ``Learning to walk in minutes using massively parallel deep reinforcement learning,'' in \emph{5th Annual Conference on Robot Learning}, 2021.

\bibitem{carpentier2019pinocchio}
J.~Carpentier, G.~Saurel, G.~Buondonno, J.~Mirabel, F.~Lamiraux, O.~Stasse, and N.~Mansard, ``The pinocchio c++ library -- a fast and flexible implementation of rigid body dynamics algorithms and their analytical derivatives,'' in \emph{IEEE International Symposium on System Integrations (SII)}, 2019.

\bibitem{ferreau2014qpOASES}
H.~Ferreau, C.~Kirches, A.~Potschka, H.~Bock, and M.~Diehl, ``{qpOASES}: A parametric active-set algorithm for quadratic programming,'' \emph{Mathematical Programming Computation}, vol.~6, no.~4, pp. 327--363, 2014.

\bibitem{todorov2012mujoco}
E.~Todorov, T.~Erez, and Y.~Tassa, ``Mujoco: A physics engine for model-based control,'' in \emph{2012 IEEE/RSJ International Conference on Intelligent Robots and Systems}.\hskip 1em plus 0.5em minus 0.4em\relax IEEE, 2012, pp. 5026--5033.

\end{thebibliography}

\end{CJK}
\end{document}